\documentclass[letterpaper]{article} 

\PassOptionsToPackage{authoryear}{natbib}

\usepackage[final]{nips_2017otml}

\usepackage[utf8]{inputenc} 
\usepackage[T1]{fontenc}    
\usepackage{hyperref}       
\usepackage{url}            
\usepackage{booktabs}       
\usepackage{amsfonts}       
\usepackage{nicefrac}       
\usepackage{microtype}      

\usepackage{tikz}       
\usetikzlibrary{matrix}  
\usepackage{graphicx} 
\usepackage{amsmath,amssymb,amsthm}
\usepackage{bm} 
\newtheorem{thm}{Theorem}[section]

\newtheorem{lem}[thm]{Lemma}

\newtheorem{dfn}{Definition}[section]

\newcommand{\refeq}[1]{\eqref{#1}} 
\newcommand{\reffig}[1]{Figure~\ref{#1}}

\newcommand{\refapp}[1]{Appendix~\ref{#1}}
\newcommand{\refsec}[1]{Section~\ref{#1}}

\newcommand{\refthm}[1]{Theorem~\ref{#1}}

\newcommand{\reflem}[1]{Lemma~\ref{#1}}
\newcommand{\xx}{\bm{x}}
\newcommand{\mmu}{\bm{\mu}}
\newcommand{\dd}{\mathrm{d}}
\newcommand{\eps}{{\bm{\varepsilon}}}
\newcommand{\RR}{\mathbb{R}}

\newcommand{\EE}{\mathbb{E}}
\newcommand{\spW}{\mathcal{P}_2}
\newcommand{\metW}{\mathfrak{g}_2}

\newcommand{\wdot}{\,\cdot\,}

\newcommand{\tr}{\mathsf{tr}\,}

\newcommand{\diag}{\mathsf{diag}\,}
\newcommand{\grad}{\mathsf{grad}\,}

\newcommand{\almost}{\mathsf{a.e.}\,}

\newcommand{\data}{\mu}
\newcommand{\noise}{\nu}
\newcommand{\dae}{\bm{\Phi}}
\newcommand{\cdae}{\bm{\varphi}}
\newcommand{\daenet}{g}

\newcommand{\ent}{\mathcal{H}}
\newcommand{\free}{\mathcal{F}}

\newcommand{\dt}{\tau}

\newcommand{\tbound}{T}

\newcommand{\flux}{\nabla V}
\newcommand{\potential}{\nabla V}
\newcommand{\gauss}{W}
\newcommand{\tmap}{\bm{g}}
\newcommand{\func}{\bm{h}}
\newcommand{\emb}[1]{\textbf{#1}}%
\title{Transportation analysis of denoising autoencoders: \\a novel method for analyzing deep neural networks}

\author{
	Sho~Sonoda\\
	School of Advanced Science and Engineering\\
	Waseda University\\
  \texttt{sho.sonoda@aoni.waseda.jp} \\
  \And
  Noboru~Murata\\
	School of Advanced Science and Engineering\\
	Waseda University\\
  \texttt{noboru.murata@eb.waseda.ac.jp} \\
}

\begin{document}

\maketitle

\begin{abstract}
The feature map obtained from the denoising autoencoder (DAE) is investigated by determining transportation dynamics of the DAE, which is a cornerstone for deep learning. Despite the rapid development in its application, deep neural networks remain analytically unexplained, because the feature maps are nested and parameters are not faithful. In this paper, we address the problem of the formulation of nested complex of parameters by regarding the feature map as a transport map. Even when a feature map has different dimensions between input and output, we can regard it as a transportation map by considering that both the input and output spaces are embedded in a common high-dimensional space. In addition, the trajectory is a geometric object and thus, is independent of parameterization. In this manner, transportation can be regarded as a universal character of deep neural networks. By determining and analyzing the transportation dynamics, we can understand the behavior of a deep neural network. In this paper, we investigate a fundamental case of deep neural networks: the DAE. We derive the transport map of the DAE, and reveal that the infinitely deep DAE transports mass to decrease a certain quantity, such as entropy, of the data distribution. These results though analytically simple, shed light on the correspondence between deep neural networks and the Wasserstein gradient flows.

\end{abstract}

\section{Introduction}
Despite the rapid development in its application, the deep structure of neural networks remains analytically unexplained
because (1) functional composition has poor compatibility with the basics of machine learning: ``basis and coefficients,''
and (2) the parameterization of neural networks is not faituful and thus parametric arguments are subject to technical difficulties such as local minima and algebraic singularities.
In this paper, we introduce the transportation interpretation of deep neural networks;
we regard a neural network with $m$-inputs and $n$-outputs as a vector-valued map $\tmap : \RR^m \to \RR^n$,
and interpret $\tmap$ as a transport map that transforms the input vector $\xx \in \RR^m$ to $\tmap(\xx) \in \RR^n$.
Because the composition of transport maps is also a transport map, a trajectory is the natural model of the composition structure of deep neural networks.
Furthermore, because a trajectory is independent of its parameterization, redundant parameterization of neural networks is avoided.
By determining and analyzing the transportation dynamics of a deep neural network, we can understand the behavior of that network.
For example, we can expect that in a deep neural network that distinguishes the pictures of \emph{dogs and cats},
the feature extractor would be a transport map that separates the input vectors of \emph{dogs and cats} apart, like 
 the physical phenomenon of 
\emph{oil and water} being immiscible.
It is noteworthy that the input and output dimensions of a feature map in a neural network rarely coincide with each other.
Nevertheless, we can regard the feature map in a neural network as a transport map by considering that both the input and output spaces are embedded in a common high-dimensional space. In this manner, we can always assign a trajectory with a deep neural network, and transportation is therefore a universal character of deep neural networks.

The denoising autoencoder (DAE)---used to obtain a good representation of data---is a cornerstone for deep learning,
or representation learning.  
The traditional autoencoder is a neural network that is trained as an identity map $\tmap(\xx) = \xx$.
The hidden layer of the network is used as a feature map, which is often called the ``code'' because, in general, the activation pattern appears random and encoded.
\citet{Vincent2008} introduced DAE as a heuristic modification of traditional autoencoders to increase robustness.
In this case, the DAE is trained as a ``denoising'' map
\begin{align*}
\tmap(\widetilde{\xx}) \approx \xx,
\end{align*}
of deliberately corrupted inputs $\widetilde{\xx}$.
Though the \emph{corrupt and denoise} principle is simple, it is successfully used for deep learning,
and has therefore, inspired many representation learning algorithms \citep{Vincent2010, Vincent2011, Rifai2011, Bengio2013, Bengio2014, Alain2014}. 
Though the term ``DAE'' is the name of a training method, as long as there is no risk of confusion,
we abbreviate ``a training result $\tmap$ of the DAE'' as ``a DAE $\tmap$''.

As discussed later, we found that when the corruption process is additive, i.e., $\widetilde{\xx}=\xx + \eps$ with some noise $\eps$,
then the DAE $\tmap$ takes the form 
\begin{align}
&\tmap_t(\widetilde{\xx}) = \widetilde{\xx} - \EE_t[\eps | \widetilde{\xx}], \label{eq:dae}
\end{align}
where $t$ denotes noise variance, and the expectation is taken with respect to a posterior distribution of noise $\eps$ given $\widetilde{\xx}$.
We can observe that the DAE \refeq{eq:dae} is composed of the traditional autoencoder $\widetilde{\xx} \mapsto \widetilde{\xx}$ and the denoising term $\widetilde{\xx} \mapsto  -\EE_t[\eps | \widetilde{\xx}]$. From a statistical viewpoint, this form is reasonable because a DAE $\tmap$ is \emph{an estimator of the mean}, or \emph{the location parameter}.
Specifically, given a corrupted input $\widetilde{\xx}$ of an unknown truth $\xx$, $\tmap(\widetilde{\xx})$ is an estimator of $\xx$.

In this study, we interpret \refeq{eq:dae} as a transport map, by regarding the denoising term as a displacement vector from the origin $\widetilde{\xx}$.
In addition, we regard the noise variance $t$ as \emph{transport time}.
As time $t$ evolves, the data distribution $\data_0$ will be deformed to $\data_t$ according to the mass transportation given by $\tmap_t$, i.e.,
$\data_t$ is the \emph{pushforward measure} of $\data_0$ by $\tmap_t$,
and is denoted by $\data_t = \tmap_{t\sharp} \data_0$.
Because $\tmap_t$ defines a time-dependent dynamical system, $\tmap_t$ is difficult to analyze.
Instead, we focus on $\data_t$, and show that $\data_t$ evolves according to a Wasserstein gradient flow with respect to a certain potential functional $\free[\data_t]$, which is independent of time. In general, a DAE is identified by $\free$.

\begin{figure}[th]
	\centering
	\includegraphics[width=0.5\textwidth]{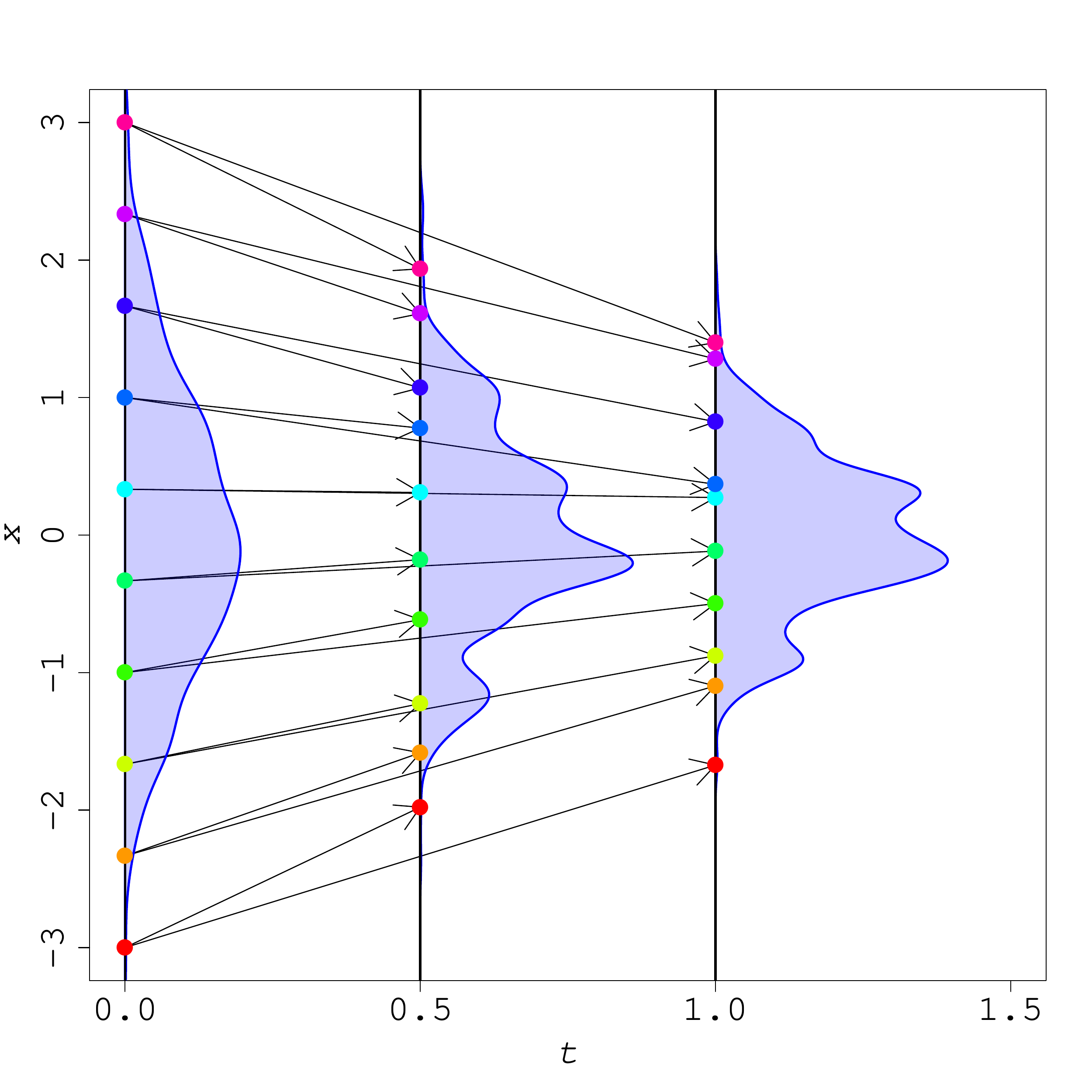}
	\vspace{-0.5cm}
	\caption{Gaussian denoising autoencoder, which is one of the most fundamental version of deep neural networks, transports mass, from the left to the right, to decrease the Shannon-Boltzmann entropy of data. The $x$-axis is the $1$-dimensional input/output space, the $t$-axis is the variance of the Gaussian noise, and $t$ is the transport time. The leftmost distribution depicts the original data distribution $\data_0 = \mathcal{N}(0,1)$. The middle and rightmost distributions depict the pushforward distribution $\data_t = \tmap_{t\sharp} \data_0$, associated with the transportation by two DAEs with noise variance $t=0.5$ and $t=1.0$, respectively. As $t$ increases, the variance of the pushforward distribution decreases.}
	\label{fig:dae1dim}
\end{figure}

In the following sections, we determine and analyze the transport map of DAEs.
In \refsec{sec:sdae}, we show that $\tmap_t$ is given by \refeq{eq:dae},
and that $\data_t$ evolves according to the \emph{continuity equation} as $t \to 0$.
Then, in \refsec{sec:ddae}, we consider the composition of DAEs, or a deep DAE,
and show that the continuum limit of the compositions satisfies the continuity equation at every time $t$.
Finally, in \refsec{sec:wgrad}, we explain the association between the DAE and the Wasserstein gradient flow.

\subsection{A minimum introduction to Wasserstein gradient flow}
The Wasserstein gradient flow \citep[\S~23]{Villani2009}, also known as the Otto calculus and the abstract gradient flow,
is an infinite-dimensional gradient flow defined on the $L^2$-Wasserstein space $\spW(\RR^m)$.
Here, $\spW(\RR^m)$ is a family of sufficiently smooth probability density functions on $\RR^m$ that have at least second moments,
equipped with $L^2$-Wasserstein metric $\metW$; $\metW$ is an infinite-dimensional Riemannian metric that is compatible with $L^2$-Wasserstein distance $W_2$; and $W_2$ is a distance between two probability densities in $\spW(\RR^m)$, which coincides with the infimum of the total Euclidean cost to transport mass that is distributed according to $\mu$ to $\nu$.
In summary, $\spW(\RR^m)$ is a functional Riemannian manifold, and 
the infinite-dimensional gradient operator $\grad$ on $\spW(\RR^m)$ is defined via metric $\metW$.

\subsection{Related works} \label{sec:relatedwork}
\citet{Alain2014} is the first to derive a special case of \refeq{eq:dae}, and their paper has been a motivation for the present study.
While we investigated a deterministic formulation of DAEs---the transport map---
they developed a probabilistic formulation of DAEs, i.e., generative modeling \citep{Alain2015}.
Presently, various formulations based on this generative modeling method are widespread; for example, variational autoencoder \citep{Kingma2014a}, minimum probability flow \citep{Sohl-Dickstein2015}, and adversarial generative networks (GANs) \citep{Goodfellow2014}.
In particular, Wasserstein GAN \citep{Arjovsky2017} employed Wasserstein geometry to reformulate and improve GANs.

\section{DAE} \label{sec:sdae}

We formulate the DAE as a variational problem, and show that the minimizer $\tmap^*$, or the training result, is a transport map.
Because a single training result of the DAE typically produces a neural network,
even though the variational formulation is independent of the choice of approximators,
we refer to the minimizer as a \emph{DAE.}
We further investigate the initial velocity vector field $\partial_t \tmap_{t=0}$ for mass transportation,
and show that the data distribution $\data_t$ evolves according to the \emph{continuity equation}.

\subsection{Training procedure of DAE}
Let $\xx$ be an $m$-dimensional random vector that is distributed according to $\data_0$,
and $\widetilde{\xx}$ be its corruption defined by
\begin{align*}
\widetilde{\xx} = \xx + \eps, \quad \eps \sim \noise_t
\end{align*}
where $\noise_t$ denotes the noise distribution parametrized by variance $t > 0$.
A basic example of $\noise_t$ is the Gaussian noise with mean $0$ and variance $t$, i.e. $\noise_t = \mathcal{N}(0,tI)$.

The DAE is a function that is trained to remove corruption $\widetilde{\xx}$ and restore it to the original $\xx$;
this is equivalent to training a function $\tmap$ for minimizing an objective function, i.e.,
\begin{align}
L[\tmap] := \EE_{\xx,\widetilde{\xx}} | \tmap( \widetilde{\xx} ) - \xx |^2. \label{eq:objective}
\end{align}
In this study, we assume that
$\tmap$ 
is a universal approximator, which need not be a neural network, and thus can attain a minimum.
Typical examples of $\tmap$ are neural networks with sufficiently large number of hidden units, $B$-splines, random forests, and kernel machines.

\subsection{Transport map of DAE}

The \emph{global} minimizer of \refeq{eq:objective} is \emph{explicitly} obtained as follows.
\begin{thm}[{Generalization of \citep[Theorem~1]{Alain2014}}] \label{thm:sonoda}
	For every $\data_0$ and $\nu_t$, $L[\tmap]$ attains the global minimum at
\begin{align} 
\tmap^*(\widetilde{\xx})
&= \EE_t[\xx | \widetilde{\xx}] 
= \frac{1}{\data_0 * \noise_t (\widetilde{\xx})} \int_{\RR^m} \xx \noise_t(\xx-\widetilde{\xx}) \data_0(\xx) \dd \xx, \label{eq:alain} \\
&= \widetilde{\xx} - \EE_t[ \eps | \widetilde{\xx} ] 
 = \widetilde{\xx} - \frac{1}{\data_0 * \noise_t(\widetilde{\xx})} \int_{\RR^m} \eps \noise_t(\eps) \data_0 (\widetilde{\xx} - \eps) \dd \eps,
\label{eq:sonoda}
\end{align}
where $*$ denotes the convolution operator.
\end{thm}

Henceforth, we refer to the minimizer as a \emph{DAE,} and symbolize \refeq{eq:sonoda} by $\dae_t$. That is,
\begin{align}
\dae_t(\xx) := \xx - \frac{1}{\data_0 * \noise_t(\xx)} \int_{\RR^m} \eps \noise_t(\eps) \data_0 (\xx - \eps) \dd \eps. \nonumber
\end{align}

As previously stated, the DAE $\dae_t(\xx)$ is composed of the identity map $\xx \mapsto \xx$ and the denoising map $\xx \mapsto -\EE_t[\eps|\xx]$.
In particular, when $t=0$, the denoising map vanishes and DAE reduces to a traditional autoencoder.
We reinterpret the DAE $\dae_t(x)$ as a \emph{transport map with transport time $t$} that transports mass at $\xx \in \RR^m$ toward $\xx + \Delta \xx \in \RR^m$ with displacement vector $\Delta \xx = -\EE_t[\eps | \xx]$.

Note that the variational calculation first appeared in
\citep[Theorem~1]{Alain2014}, in which the authors obtained \refeq{eq:alain}.
In statistics, \refeq{eq:sonoda} is known as Brown's representation of the posterior mean \citep{George2006}.
This is not just a coincidence because, as \refeq{eq:alain} suggests, the DAE is an estimator of the mean.  

\subsection{Initial velocity of the transport map}
For the sake of simplicity and generality, we consider a generalized form of short time transport maps:
\begin{align}
\tmap_t(\xx) := \xx + t \flux_t(\xx), \label{eq:tmap}
\end{align}
with some potential function $V_t$, and the potential velocity field, or flux, $\flux_t$.
For example, as shown in \refeq{eq:gdae.sonoda}, the Gaussian DAE is expressed in this form.
Note that the establishment of a reasonable correspondence between $\dae_t$ and $\flux_t$ for an arbitrary $\noise_t$
 is an \emph{open question}.

For the initial moment ($t=0$), the following lemma holds.
\begin{lem} \label{lem:contieq}
Given a data distribution $\data_0$, 
the pushforward measure $\data_t := \tmap_{t\sharp} \data_0$ satisfies the continuity equation
\begin{align}
\partial_t \data_{t}(\xx) = - \nabla \cdot [ \data_{t}(\xx) \flux_{t}(\xx) ], \quad t=0 \label{eq:dae.contieq}
\end{align}
where $\nabla \cdot $ denotes the divergence operator on $\RR^m$.
\end{lem}
The proof is given in \refapp{proof:contieq}.
Intuitively, the statement seems natural because \eqref{eq:tmap} is a standard setup for the continuity equation.
Note that this relation does not hold in general. Particularly, $\partial_t \data_{t} \neq - \nabla \cdot [ \data_{t} \flux_{t} ]$ for $t>0$.
This is because time-dependent dynamics should be written as an ordinary differential equation such as $\partial_t \tmap_t(\xx) = \flux_t (\xx)$.

\subsection{Example: Gaussian DAE}
When $\noise_t = \mathcal{N}(0, t I)$, 
the posterior mean $\EE_t[\eps | \xx]$ is analytically obtained as follows.
\begin{align*}
\EE_t[\eps | \xx]
= - \frac{t \nabla \noise_t * \data_0(\xx)}{\noise_t * \data_0(\xx)}
= -t \nabla \log [ \noise_t * \data_0 (\xx)],
\end{align*}
where the first equation follows by Stein's identity
\begin{align*}
-t \nabla \noise_t(\eps) = \eps \, \noise_t(\eps),
\end{align*}
which is known to hold only for Gaussians.
\begin{thm} Gaussian DAE $\dae_t$ is given by
\begin{align} 
\dae_t(\xx) = \xx + t \nabla \log[ \gauss_t * \data_0](\xx),  \label{eq:gdae.sonoda}
\end{align}
with Gaussian $\gauss_t(\xx) := (2 \pi t)^{-m/2} \exp \left( -|\xx|^2/2t \right)$.
\end{thm}

When $t \to 0$, the initial velocity vector is given by the \emph{score} (i.e., score matching)
\begin{align}
\partial_t \dae_{t=0} (\xx) = \lim_{t \to 0} \frac{\dae_t(\xx) - \xx}{t} = \nabla \log \data_0 (\xx). \label{eq:dae.tp.init}
\end{align}
Hence, by substituting the score \refeq{eq:dae.tp.init} in the continuity equation \refeq{eq:dae.contieq}, we have
\begin{align*}
\partial_t \data_{t=0}(\xx)
&= -\nabla \cdot [ \data_{0} (\xx) \nabla \log \data_0(\xx) ]\\
&= -\nabla \cdot [ \nabla \data_0(\xx) ]\\
&= -\triangle \data_0(\xx),
\end{align*}
where $\triangle$ denotes the Laplacian on $\RR^m$.
\begin{thm} \label{thm:dae.pfinit}
	The pushforward measure of Gaussian DAE satisfies the \emph{backward heat equation}:
	\begin{align}
	\partial_t \data_{t=0}(\xx) = -\triangle \data_0(\xx). \label{eq:dae.pf.init}
	\end{align}
\end{thm}
We shall investigate the backward heat equation in \refsec{sec:wgrad}.

\section{Deep Gaussian DAEs} \label{sec:ddae}
As a concrete example of deep DAEs, we investigate further the Gaussian DAE ($\noise_t = \mathcal{N}(0,tI)$).
We introduce the composition of DAEs, and the continuous DAE as an infinitesimal limit.
We can understand the composition of DAEs as the \emph{Eulerian broken line approximation} of a continuous DAE.

\subsection{Composition of Gaussian DAEs}
Let $\xx_0$ be an $m$-dimensional input vector that is subject to data distribution $\data_0$,
and $\dae_0 : \RR^m \to \RR^m$ be a DAE that is trained for $\data_0$ with noise variance $\tau_0$.
Write $\xx_1 := \dae_0(\xx_0)$.
Then $\xx_1$ is a random vector in $\RR^m$ that is subject to the pushforward measure $\data_1 := \dae_{0 \sharp} \data_0$,
and thus, we can train another DAE $\dae_1 : \RR^m \to \RR^m$ using $\data_1$ with noise variance $\tau_1$.
By repeating the procedure, we can obtain $\xx_{\ell+1} := \dae_\ell(\xx_\ell) \sim \data_{\ell+1} := \dae_{\ell \sharp} \data_\ell$ from $\xx_\ell \sim \data_\ell$, and $\dae_{\ell+1}$ with variance $\tau_{\ell + 1}$.
We write the composition of DAEs by
\begin{align*}
\dae_{0:L}^t(\xx) := \dae_L \circ \cdots \circ \dae_0(\xx),
\end{align*}
where $t$ denotes ``total time''; $t := \tau_0 + \cdots + \tau_L$.
By definition, at every $t_\ell := \tau_0 + \cdots + \tau_\ell$, the velocity vector of a composition of DAEs coincides with
the score
\begin{align*}
\partial_{t} \dae_{0:\ell}^{t=t_\ell}(\xx) = \nabla \log \data_{t_\ell}(\xx).
\end{align*}

\subsection{Continuous Gaussian DAE}

We set total time $t=\tau_0 + \cdots + \tau_L$ and take limit $L \to \infty$ of the layer number.
Then, we can see that the velocity vector of ``infinite composition of DAEs'' $\lim_{L \to \infty} \dae_{0:L}^t$ tends to coincide with the continuity equation at every time.
Hence, we introduce an ideal version of DAE as follows.
\begin{dfn}
	Set data distribution $\data_0 \in \spW(\RR^m)$.
	We call the solution operator, or flow $\cdae_t : \RR^m \to \RR^m$, of the following dynamics as the \emph{continuous DAE}.
	\begin{align}
	\frac{\dd }{\dd t} \xx(t) = \nabla \log \data_t( \xx(t) ), \quad t \geq 0 \label{eq:dynamics}
	\end{align}
	where $\data_t := \cdae_{t\sharp} \data_0$.
\end{dfn}
The limit converges to a continuous DAE when, for example, the score $\nabla \log \data_t$ is Lipschitz continuous at every time $t$,
because trajectory $t \mapsto \dae_{0:L}^t(x_0)$ corresponds to a Eulerian broken line approximation of the integral curve $t \mapsto \cdae_t(x)$ of \eqref{eq:dynamics}.

The following property is immediate from \refthm{thm:dae.pfinit}.
\begin{thm} \label{thm:cdae.backward}
	Let $\cdae_t$ be a continuous DAE trained for $\data_0 \in \spW(\RR^m)$.
	Then, the pushforward measure $\data_t := \cdae_{t \sharp} \data_0$ is the solution of the initial value problem
	\begin{align}
	\partial_t \data_t(\xx) = - \triangle \data_t(\xx), \quad \data_{t=0}(\xx) =\data_0(\xx) \label{eq:backward}
	\end{align}
	which we refer to as the \emph{backward heat equation}.
\end{thm}

The backward heat equation \refeq{eq:backward} is equivalent to the following \emph{final value problem} for the ordinary heat equation:
\begin{align*}
\partial_t u_t(\xx) = \triangle u_t(\xx), \quad u_{t = \tbound}(\xx) = \data_0(\xx) \quad \mbox{ for some } \tbound
\end{align*}
where $u_t$ denotes a probability measure on $\RR^m$.
Indeed,
\begin{align*}
\data_t(\xx) = u_{\tbound - t}(\xx),
\end{align*}
is the solution of \refeq{eq:backward}.
In other words, backward heat equation describes the time reversal of an ordinary diffusion process.

\subsection{Numerical example of trajectories}
\begin{figure}[h]
	\centering
	\begin{tabular}{cc}
		\begin{minipage}{0.4\hsize}
			\centering
			\includegraphics[width=\textwidth]{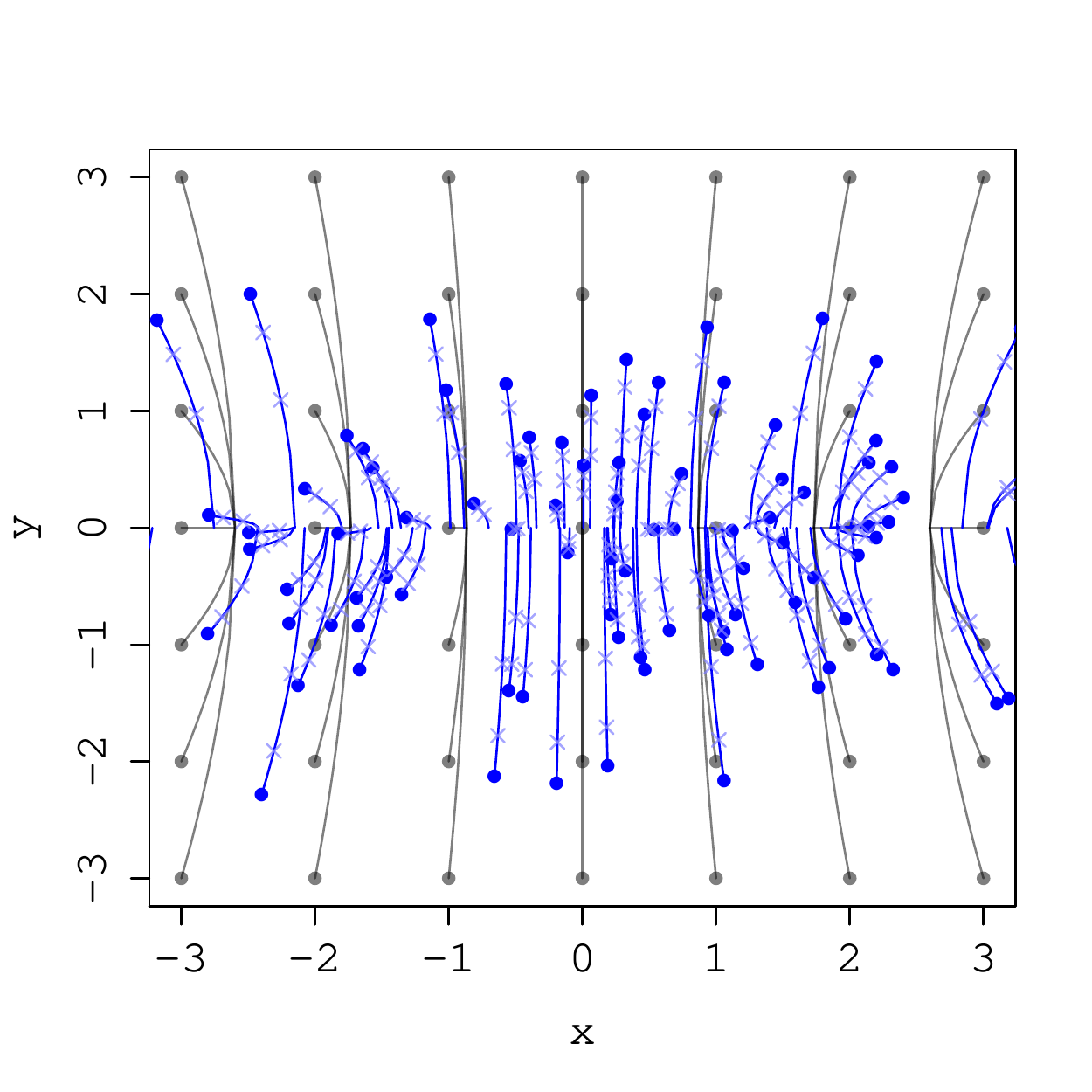}
\vspace{-0.4cm}
		\end{minipage} &
		\begin{minipage}{0.4\hsize}
			\centering
			\includegraphics[width=\textwidth]{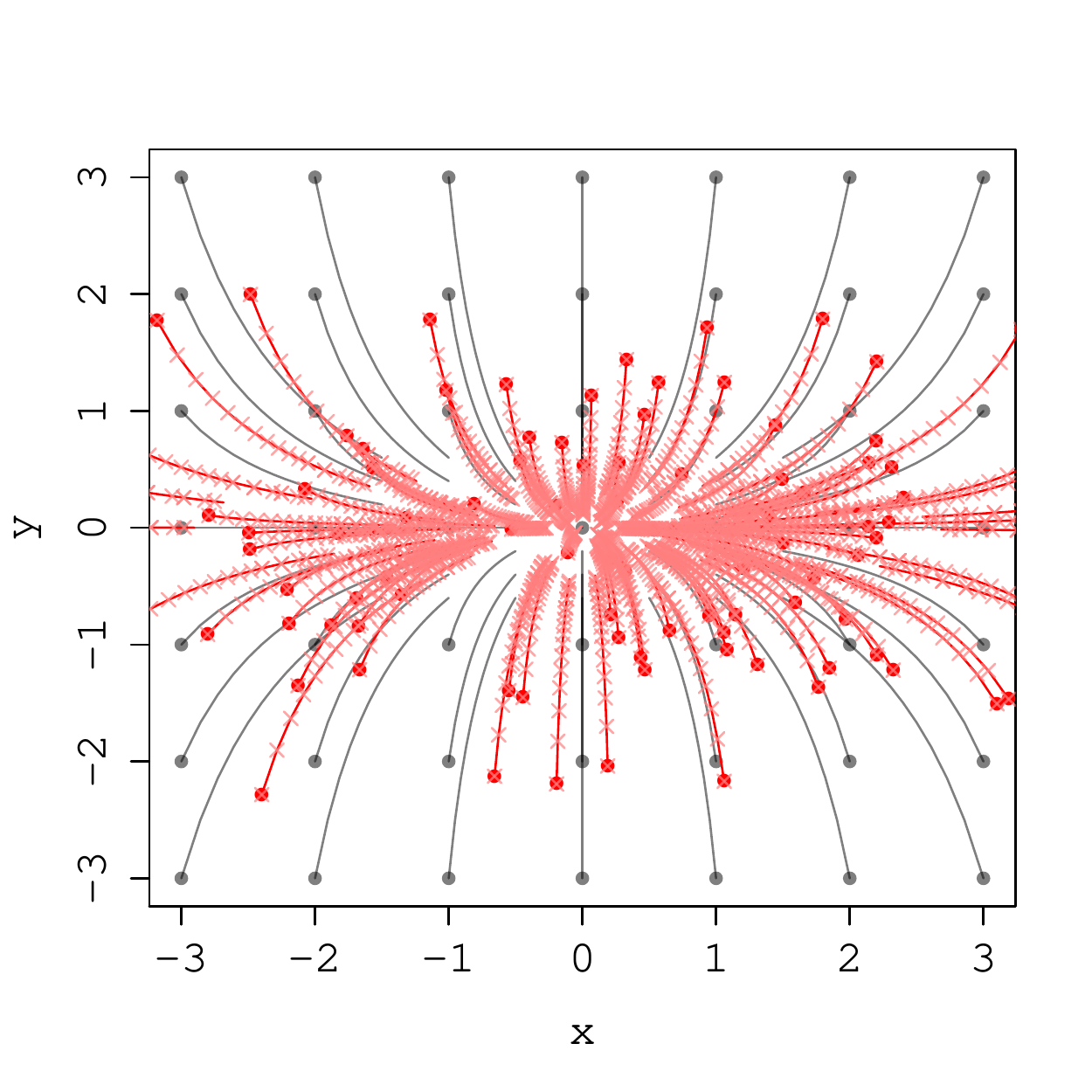}
\vspace{-0.4cm}
		\end{minipage} \\
		\begin{minipage}{0.4\hsize}
			\centering
			\includegraphics[width=\textwidth]{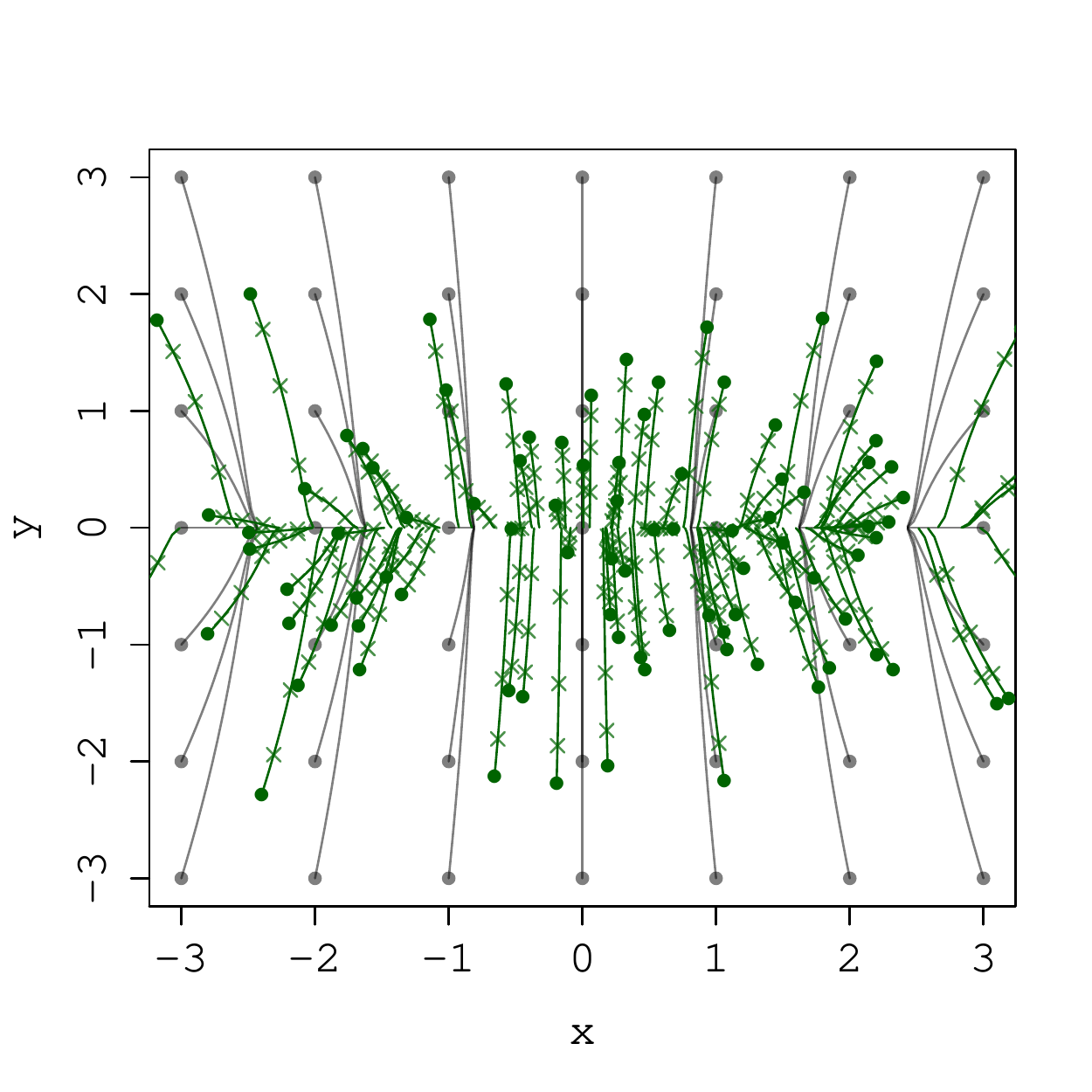}
\vspace{-0.4cm}
		\end{minipage} &
		\begin{minipage}{0.4\hsize}
			\centering
			\includegraphics[width=\textwidth]{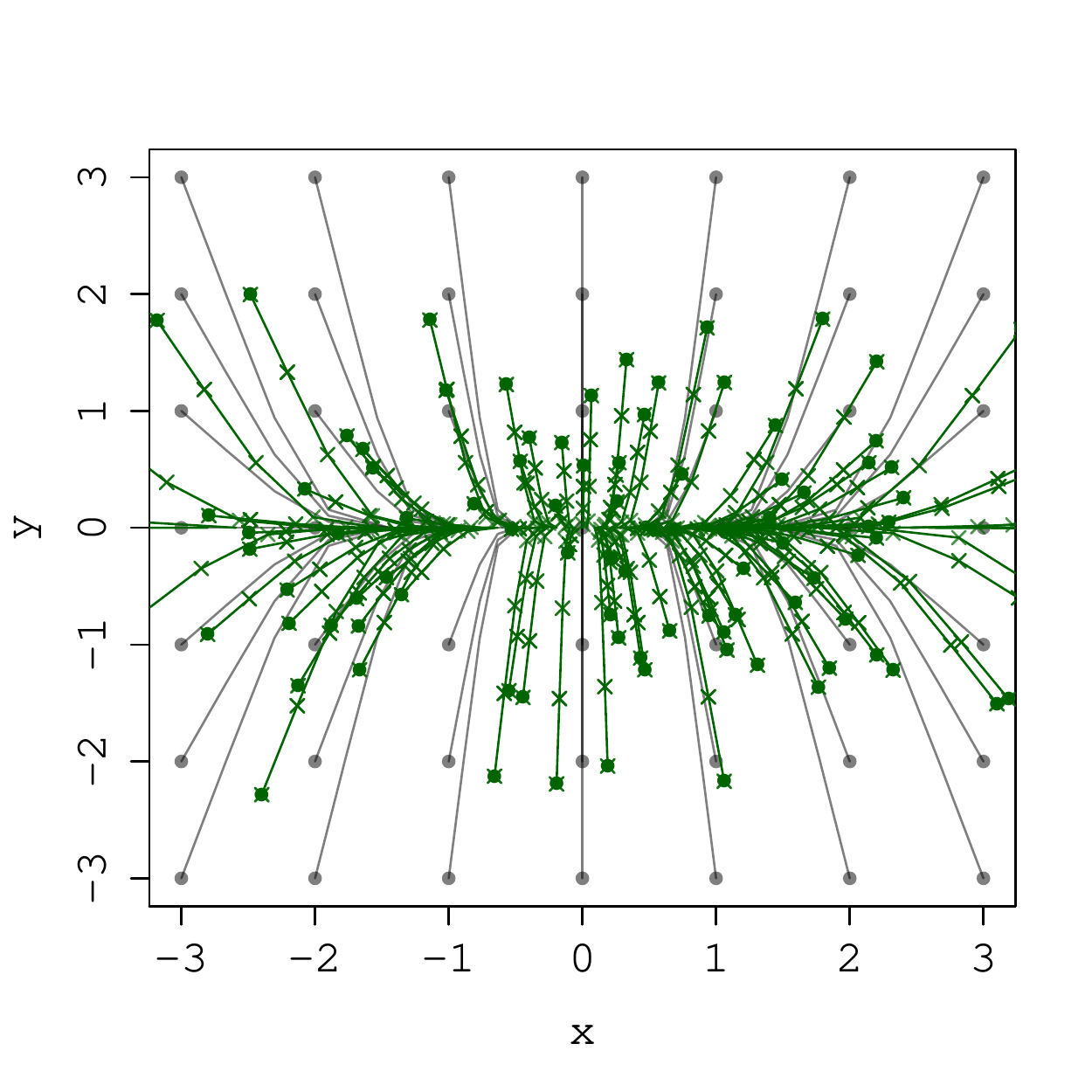}
\vspace{-0.4cm}
		\end{minipage} \\
	\end{tabular}
\vspace{-0.4cm}
	\caption{
		Trajectories of DAEs trained for the same data distribution $\data_0 = \mathcal{N}([0,0],\diag[2,1])$.
		\emb{Top Left}: continuous DAE $t \mapsto \cdae_t$.
		\emb{Top Right}: DAE $t \mapsto \dae_t$.
		\emb{Bottom Left}: compositions of DAEs  $t \mapsto \dae_{0:L}^t$ with $\dt = 0.05$.
		\emb{Bottom Right}: compositions of DAEs  $t \mapsto \dae_{0:L}^t$ with $\dt = 0.5$.
		\emb{Gray lines} start from the regular grid. \emb{Colored lines} start from the samples drawn from $\data_0$.
		\emb{Midpoints} are plotted every $\dt = 0.2$.}
	\label{fig:dae.cdae.comdae}
\end{figure}

\reffig{fig:dae.cdae.comdae} compares the trajectories of four DAEs trained for the same data distribution
\begin{align*}
\data_0 = \mathcal{N} \left( [0,0],\diag[2,1] \right).
\end{align*}
The trajectories are analytically calculated as 
\begin{align}
\cdae_t(\xx) &= \sqrt{ I - 2 t \Sigma_0^{-1 }}(\xx - \mmu_0) + \mmu_0, \label{eq:ex.cdae}
\end{align}
and
\begin{align}
\dae_{t}(\xx) &= (I + t \Sigma_0^{-1})^{-1}\xx + (I + t^{-1}\Sigma_0)^{-1} \mmu_0, \label{eq:ex.dae}
\end{align}
where $\mu_0$ and $\Sigma_0$ are mean and covariance matrix of the normal distribution, respectively.

The continuous DAE \refeq{eq:ex.cdae} attains the singularity at $t = 1/2$.
On the contrary, the DAE \refeq{eq:ex.dae} slows down as $t \to \infty$ and never attains the singularity in finite time.
As $L$ tends to infinity, $\dae_{0:L}^t$ draws a similar orbit as the continuous DAE $\cdae_t$; the curvature of orbits also changes according to $\dt$.

\section{Wasserstein gradient flow} \label{sec:wgrad}

As an analogy of the Gaussian DAE,
we can expect that the pushforward measure $\data_t := \cdae_{t \sharp} \data_0$ of
a general continuous DAE $\cdae_t$ satisfies the continuity equation:
\begin{align}
\partial_t \data_{t}(\xx) = - \nabla \cdot [ \data_{t}(\xx) \flux_{t}(\xx) ], \quad t>0. \label{eq:conti}
\end{align}

According to Otto calculus \citep[Ex.15.10]{Villani2009},
the solution $\data_t$ coincides with a trajectory of the \emph{Wasserstein gradient flow}
\begin{align}
\frac{\dd}{\dd t} \data_t = - \grad \free[\data_t], \label{eq:gflow}
\end{align}
with respect to a potential functional $\free : \spW(\RR^m) \to \RR$.
Here, $\grad$ denotes the gradient operator on $L^2$-Wasserstein space $\spW(\RR^m)$,
and $\free$ satisfies the following equation:
\begin{align*}
\frac{\dd }{\dd t} \free[\data_t] = \int_{\RR^m} \potential_t (\xx) [\partial_t \data_t] (\xx) \dd x.
\end{align*}

Recall that the $L^2$-Wasserstein space $\spW(\RR^m)$ is a functional manifold.
While \refeq{eq:gflow} is an ordinary differential equation on the space $\spW(\RR^m)$ of probability density functions,
\refeq{eq:conti} is a partial differential equation on the Euclidean space $\RR^m$. Hence, we use different notations for the time derivatives: $\frac{\dd}{\dd t}$ and $\partial_t$.

The Wasserstein gradient flow \refeq{eq:gflow} possesses
a distinct advantage that the potential functional $\free$ does not depend on time $t$.
In the following subsections, we will see both the Boltzmann entropy and the Renyi entropy as examples of $\free$.

\subsection{Example: Gaussian DAE}
According to Wasserstein geometry, an ordinary heat equation corresponds to a Wasserstein gradient flow that \emph{increases} the entropy functional $\ent[\data] := -\int \data(\xx) \log \data(\xx) \dd \xx$ \citep[Th.~23.19]{Villani2009}.
Consequently, we can conclude that the feature map of the Gaussian DAE is a transport map that \emph{decreases} the entropy  of the data distribution:
\begin{align}
\frac{\dd}{\dd t} \data_t = - \grad \ent[\data_t], \quad \data_{t=0} = \data_0.
\end{align}
This is immediate, because when $\free=\ent$,
then $V(\xx) = -\log \data_t (\xx)$; thus,
\begin{align*}
\grad \ent[\data_t]
= \nabla \cdot [ \data_t \nabla \log \data_t ]
= \nabla \cdot \left[ \data_t \frac{\nabla \data_t}{\data_t} \right]
= \triangle \data_t,
\end{align*}
which means \refeq{eq:conti} reduces to the backward heat equation.

\subsection{Example: Renyi Entropy}
Similarly, when $\free$ is the Renyi entropy
\begin{align*}
\ent^\alpha[\data] := \int_{\RR^m} \frac{\data^\alpha(\xx) - \data(\xx)}{\alpha - 1} \dd \xx,
\end{align*}
then $\grad \ent^\alpha [\data_t] = \triangle \data_t^\alpha$
(see \citep[Ex.15.6]{Villani2009} for the proof) and thus \refeq{eq:conti} reduces to the \emph{backward porous medium equation}
\begin{align}
\partial_t \data_t(\xx) = - \triangle \data_t^\alpha(\xx).
\end{align}

\subsection{Numerical example of abstract trajectories}
\begin{figure}[h]
	\centering
	\includegraphics[width=0.5\textwidth]{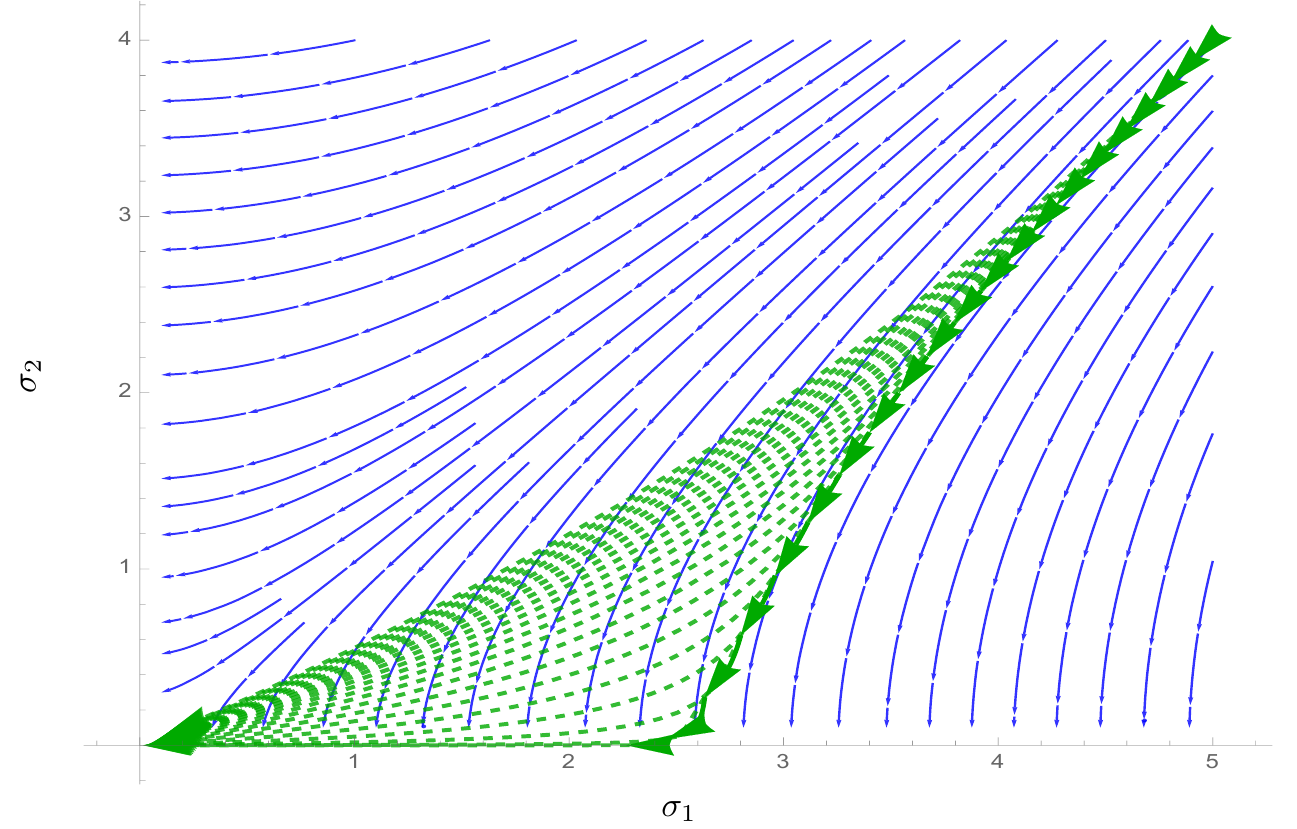}
	\caption{
		Abstract trajectories of pushforward measures in the space of bivariate Gaussians $\mathcal{N}([0,0], \diag[ \sigma^2_1, \sigma^2_2 ])$
		with entropy $\log \sigma_1 + \log \sigma_2.$
		Continuous DAE $t \mapsto \cdae_{t \sharp} \data_0$ (\emb{blue}) coincides with a Wasserstein gradient flow with respect to entropy.
	 DAE $t \mapsto \dae_{t \sharp} \data_0$ (\emb{dashed, green}) and composition of DAEs $t \mapsto \dae_{0:L \sharp}^t \data_0$ (\emb{real, green}) gradually leave the flow.
	}
	\label{fig:aflow}
\end{figure}

\reffig{fig:aflow} compares the abstract trajectories of pushforward measures in the space of bivariate Gaussians
\begin{align*}
\data_0 = \mathcal{N}([0,0], \diag[ \sigma^2_1, \sigma^2_2 ]).
\end{align*}
The entropy functional is given by
\begin{align*}
\ent(\sigma_1, \sigma_2) &= (1/2)\log | \diag[ \sigma^2_1, \sigma^2_2] | + const. \\
&= \log \sigma_1 + \log \sigma_2 + const.
\end{align*}
Note that the parameterization is reasonable, because, in this space, the Wasserstein distance between two points $(\sigma_1, \sigma_2)$ and $(\tau_1,\tau_2)$ is given by $\sqrt{(\sigma_1 - \tau_1)^2 + (\sigma_2 - \tau_2)^2}$.
The pushforward measures are analytically calculated as 
\begin{align*}
\cdae_{t \sharp} \mathcal{N}(\mmu_0,\Sigma_0) &= \mathcal{N} \left( \mmu_0, \Sigma_0 - 2 t I \right), 
\end{align*}
and
\begin{align*}
\dae_{t \sharp} \mathcal{N}( \mmu_0, \Sigma_0 ) &= \mathcal{N} \left( \mmu_0, \Sigma_0(I + t \Sigma_0^{-1})^{-2} \right),
\end{align*}
where $\mmu_0$ and $\Sigma_0$ are mean and covariance matrix of the normal distribution, respectively.

\section{Discussion}
We investigated deep denoising autoencoders (DAEs) using transportation theory.

The training algorithm of the DAE is equivalent to the minimization of $\EE_{\widetilde{\xx},\xx}|\xx - \daenet(\widetilde{\xx})|^2$ with respect to $\daenet$.
We found that the minimizer is given by a transport map \refeq{eq:sonoda}.
The initial velocity vector $\partial_t \dae_{t=0}$ of the mass transportation is given by the score $\nabla \log \data_0$.
Consequently, for Gaussian DAEs, the initial velocity $\partial_t \data_{t=0}$ of the pushforward measure coincides with the negative Laplacian $- \triangle \data_0$.
In particular, the DAE transports mass to restore the diffusion. From a statistical viewpoint, it is a natural consequence because the DAE is an estimator of the mean.

These properties are limited to $t=0$ for the DAE. Hence, we introduced the composition of DAEs and its limit, i.e., the continuous DAE.
We can understand the composition of DAEs as a Eulerian broken line approximation of a continuous DAE.
The pushforward measure $\cdae_{t \sharp} \data_0$ of the continuous Gaussian DAE satisfies the backward heat equation (\refthm{thm:cdae.backward}).
According to Wasserstein geometry, the continuous Gaussian DAE, which is an infinitely deep DAE, transports mass to decrease the entropy of the data distribution.

In general, the estimation of the time reversal of a diffusion process is an inverse problem.
In fact, our preliminary experiments indicated that the training result is sensitive to the small perturbation of training data.
However, as previously mentioned, from a statistical viewpoint, this was expected, because, by definition, a DAE is an estimator of the mean. Therefore, like a good estimator that reduces uncertainty of a parameter, the DAE will decrease entropy of the parameter.

We expect that not only the DAE, but also a wide range of deep neural networks, including both supervised and unsupervised ones, can be uniformly regarded as transport maps. For example, it is not difficult to imagine that DAEs with non-Gaussian noise correspond to other Lyapunov functionals such as the Renyi entropy and the Bregman divergence.
The form $\xx \mapsto \xx + \tmap(\xx)$ of transport maps emerges not only in DAEs, but also, for example, in ResNet \citep{He2015}.
Transportation analysis of these deep neural networks will be part of our future works.



\appendix

\section{Proof of \refthm{thm:sonoda}} \label{proof:sonoda}

	This proof follows from a variational calculation.
	Rewrite 
	\begin{align*}
	L[\tmap]
	&= \int_{\RR^m} \EE_{\eps}| \tmap(\xx +\eps) - \xx |^2 \data_0(\xx) \dd \xx\\
	&= \int_{\RR^m} \EE_{\eps}[ | \tmap(\xx') - \xx'+\eps |^2 \data_0(\xx'-\eps)] \dd \xx', \quad \xx' \gets \xx+\eps.
	\end{align*}
	Then, for every function $\func$, variation $\delta L [\func]$ is given by the directional derivative along $\func$:
	\begin{align*}
	\delta L[\func] &=\frac{d}{d s} L[\tmap + s \func] \Big|_{s=0}\\
	&=  \int_{\RR^m} \frac{\partial}{\partial s} \EE_{\eps}[| \tmap(\xx) + s \func(\xx) - \xx+\eps |^2 \data_0(\xx-\eps)] \dd \xx \Big|_{s=0} \\
	&=  2 \int_{\RR^m} \EE_{\eps}[( \tmap(\xx) - \xx+\eps ) \data_0(\xx-\eps)] \func(\xx)\dd \xx.
	\end{align*}
	At a critical point $\tmap^*$ of $L$, $\delta L[\func] \equiv 0$ for every $\func$. Hence
	\begin{align*}
	\EE_{\eps}[( \tmap^*(\xx) - \xx+\eps ) \data_0(\xx-\eps) ] =0, \quad \almost \xx,
	\end{align*}
	and we have
	\begin{align*}
	\tmap^*(\xx)
	&= \frac{\EE_{\eps}[(\xx-\eps ) \data_0(\xx-\eps)]}{\EE_\eps[\data_0(\xx-\eps)]} = \refeq{eq:alain}\\
	&= \xx - \frac{\EE_{\eps}[\eps  \data_0(\xx-\eps)]}{\EE_\eps[\data_0(\xx-\eps)]} = \refeq{eq:sonoda}. 
	\end{align*}
	The $\tmap^*$ attains the global minimum, because, for every function $h$, 
	\begin{align*}
	L[\tmap^* + \func]
	&= \int_{\RR^m} \EE_{\eps}[ | \eps - \EE_t[\eps|\xx] +\func(\xx) |^2 \data_0(\xx-\eps)] \dd \xx\\
	&=
	\int_{\RR^m} \EE_{\eps}[ | \eps - \EE_t[\eps|x] |^2 \data_0(\xx-\eps)] \dd \xx
	+ \int_{\RR^m} \EE_{\eps}[ | \func(\xx) |^2 \data_0(\xx-\eps)] \dd \xx \\ &\qquad + 2 \int_{\RR^m} \EE_{\eps}[ ( \eps - \EE_t[\eps|\xx] )  \data_0(\xx-\eps)] \func(\xx) \dd \xx\\
	&= L[\tmap^*] + L[\func] + 2 \cdot 0 \geq L[\tmap^*].
	\end{align*}

\section{Proof of \reflem{lem:contieq}} \label{proof:contieq}
	To facilitate visualization, we write $\tmap(\xx,t), \flux(\xx,t) $ and $ \data(\xx,t)$ instead of $\tmap_t(\xx), \flux_t(\xx)$, and $\data_t(\xx)$, respectively.
It immediately follows then, 
\begin{align*}
\tmap(\xx,0) = \xx, \quad \partial_t \tmap(\xx,0) = \flux(\xx,0), \quad \nabla \tmap(\xx,0) = I.
\end{align*}
According to the change of variables formula,
\begin{align*}
\data(\tmap(\xx,t),t) \cdot |\nabla \tmap(\xx,t)| = \data(\xx,0).
\end{align*}
where $| \wdot |$ denotes the determinant.
	
Take logarithm on both sides, and then differentiate with respect to $t$. Then, the RHS vanishes, and the LHS is calculated as follows.
\begin{align*}
\partial_t \log[ \data(\tmap(\xx,t),t) \cdot |\nabla \tmap(\xx,t)|]
&= \frac{\partial_t [\data(\tmap(\xx,t),t)]}{\data(\tmap(\xx,t),t)} + \partial_t \log |\nabla \tmap(\xx,t)|\\
&= \frac{(\nabla \data)(\tmap(\xx,t),t) \cdot \partial_t \tmap(\xx,t) + (\partial_t \data)(\tmap(\xx,t),t)}{\data(\tmap(\xx,t),t)} \\ & \qquad + \tr [ (\nabla \tmap(\xx,t))^{-1} \nabla \partial_t \tmap(\xx,t)]
\end{align*}
where the second term follows a differentiation formula \citep[(43)]{MatCookbook}
\begin{align*}
	\partial \log |J| = \tr[J^{-1} \partial J].
\end{align*}
	
Substitute $t \gets 0$. Then, we have
\begin{align*}
\frac{\nabla \data(\xx,0) \cdot \flux(\xx,0) + (\partial_t \data)(\xx,0)}{\data(\xx,0)} + \tr [ \nabla \flux(\xx,0) ] = 0,
\end{align*}
which leads to
\begin{align*}
\partial_t \data(\xx,0) = - \nabla \cdot [\data(\xx,0) \flux(\xx)]. 
\end{align*}

\small

\bibliographystyle{plainnat}
\bibliography{./library_short}

\begin{thebibliography}{16}
\providecommand{\natexlab}[1]{#1}
\providecommand{\url}[1]{\texttt{#1}}
\expandafter\ifx\csname urlstyle\endcsname\relax
  \providecommand{\doi}[1]{doi: #1}\else
  \providecommand{\doi}{doi: \begingroup \urlstyle{rm}\Url}\fi

\bibitem[Alain and Bengio(2014)]{Alain2014}
Guillaume Alain and Yoshua Bengio.
\newblock \href{http://jmlr.org/papers/volume15/alain14a/alain14a.pdf}{{What
  Regularized Auto-Encoders Learn from the Data Generating Distribution}}.
\newblock \emph{JMLR}, pages 3743--3773, 2014.

\bibitem[Alain et~al.(2016)Alain, Bengio, Yao, Yosinski, Thibodeau-Laufer,
  Zhang, and Vincent]{Alain2015}
Guillaume Alain, Yoshua Bengio, Li~Yao, Jason Yosinski, Eric Thibodeau-Laufer,
  Saizheng Zhang, and Pascal Vincent.
\newblock \href{http://doi.org/10.1093/imaiai/iaw003}{{GSNs : Generative
  Stochastic Networks}}.
\newblock \emph{Information and Inference}, \penalty0 (2):\penalty0 210--249,
  2016.

\bibitem[Arjovsky et~al.(2017)Arjovsky, Chintala, and Bottou]{Arjovsky2017}
Martin Arjovsky, Soumith Chintala, and L{\'{e}}on Bottou.
\newblock {Wasserstein GAN}.
\newblock Technical report, 2017.

\bibitem[Bengio et~al.(2013)Bengio, Yao, Alain, and Vincent]{Bengio2013}
Yoshua Bengio, Li~Yao, Guillaume Alain, and Pascal Vincent.
\newblock
  \href{http://papers.nips.cc/paper/5023-generalized-denoising-auto-encoders-as-generative-models.pdf}{{Generalized
  denoising auto-encoders as generative models}}.
\newblock In \emph{NIPS2013}, pages 899--907, 2013.

\bibitem[Bengio et~al.(2014)Bengio, Thibodeau-Laufer, Alain, and
  Yosinski]{Bengio2014}
Yoshua Bengio, {\'{E}}ric Thibodeau-Laufer, Guillaume Alain, and Jason
  Yosinski.
\newblock \href{http://jmlr.org/proceedings/papers/v32/bengio14.pdf}{{Deep
  Generative Stochastic Networks Trainable by Backprop}}.
\newblock In \emph{ICML2014}, pages 226--234, 2014.

\bibitem[George et~al.(2006)George, Liang, and Xu]{George2006}
Edward~I. George, Feng Liang, and Xinyi Xu.
\newblock \href{http://doi.org/10.1214/009053606000000155}{{Improved minimax
  predictive densities under Kullback-Leibler loss}}.
\newblock \emph{Annals of Statistics}, 34\penalty0 (1):\penalty0 78--91, 2006.

\bibitem[Goodfellow et~al.(2014)Goodfellow, Pouget-Abadie, Mirza, Xu,
  Warde-Farley, Ozair, Courville, and Bengio]{Goodfellow2014}
Ian Goodfellow, Jean Pouget-Abadie, Mehdi Mirza, Bing Xu, David Warde-Farley,
  Sherjil Ozair, Aaron Courville, and Yoshua Bengio.
\newblock
  \href{http://papers.nips.cc/paper/5423-generative-adversarial-nets.pdf}{{Generative
  Adversarial Nets}}.
\newblock In \emph{NIPS2014}, pages 2672--2680, 2014.

\bibitem[He et~al.(2016)He, Zhang, Ren, and Sun]{He2015}
Kaiming He, Xiangyu Zhang, Shaoqing Ren, and Jian Sun.
\newblock
  \href{http://www.cv-foundation.org/openaccess/content_cvpr_2016/papers/He_Deep_Residual_Learning_CVPR_2016_paper.pdf}{{Deep
  Residual Learning for Image Recognition}}.
\newblock In \emph{The IEEE Conference on Computer Vision and Pattern
  Recognition (CVPR)}, pages 770--778, 2016.

\bibitem[Kingma and Welling(2014)]{Kingma2014a}
Diederik~P. Kingma and Max Welling.
\newblock \href{http://arxiv.org/abs/1312.6114}{{Auto-Encoding Variational
  Bayes}}.
\newblock In \emph{ICLR2014}, pages 1--14, 2014.

\bibitem[Petersen and Pedersen(2012)]{MatCookbook}
Kaare~Brandt Petersen and Michael~Syskind Pedersen.
\newblock \href{http://www2.imm.dtu.dk/pubdb/p.php?3274}{{The Matrix Cookbook,
  Version: November 15, 2012}}.
\newblock Technical report, Technical University of Denmark, 2012.

\bibitem[Rifai et~al.(2011)Rifai, Vincent, Muller, Glorot, and
  Bengio]{Rifai2011}
Salah Rifai, Pascal Vincent, Xavier Muller, Xavier Glorot, and Yoshua Bengio.
\newblock \href{http://www.icml-2011.org/papers/455_icmlpaper.pdf}{{Contractive
  auto-encoders: explicit invariance during feature extraction}}.
\newblock In \emph{ICML2011}, pages 833--840, 2011.

\bibitem[Sohl-Dickstein et~al.(2015)Sohl-Dickstein, Weiss, Maheswaranathan, and
  Ganguli]{Sohl-Dickstein2015}
Jascha Sohl-Dickstein, Eric Weiss, Niru Maheswaranathan, and Surya Ganguli.
\newblock
  \href{http://jmlr.org/proceedings/papers/v37/sohl-dickstein15.pdf}{{Deep
  Unsupervised Learning using Nonequilibrium Thermodynamics}}.
\newblock In \emph{ICML2015}, pages 2256--2265, 2015.

\bibitem[Villani(2009)]{Villani2009}
C{\'{e}}dric Villani.
\newblock \emph{\href{http://doi.org/10.1007/978-3-540-71050-9}{{Optimal
  Transport: Old and New}}}.
\newblock Springer-Verlag Berlin Heidelberg, 2009.

\bibitem[Vincent(2011)]{Vincent2011}
Pascal Vincent.
\newblock \href{http://doi.org/10.1162/NECO_a_00142}{{A connection between
  score matching and denoising autoencoders}}.
\newblock \emph{Neural Computation}, 23\penalty0 (7):\penalty0 1661--1674,
  2011.

\bibitem[Vincent et~al.(2008)Vincent, Larochelle, Bengio, and
  Manzagol]{Vincent2008}
Pascal Vincent, Hugo Larochelle, Yoshua Bengio, and Pierre-Antoine Manzagol.
\newblock
  \href{http://www.machinelearning.org/archive/icml2008/papers/592.pdf}{{Extracting
  and Composing Robust Features with Denoising Autoencoders}}.
\newblock In \emph{ICML2008}, pages 1096--1103, 2008.

\bibitem[Vincent et~al.(2010)Vincent, Larochelle, Lajoie, Bengio, and
  Manzagol]{Vincent2010}
Pascal Vincent, Hugo Larochelle, Isabelle Lajoie, Yoshua Bengio, and
  Pierre-Antoine Manzagol.
\newblock
  \href{http://www.jmlr.org/papers/volume11/vincent10a/vincent10a.pdf}{{Stacked
  denoising autoencoders: learning useful representations in a deep network
  with a local denoising criterion}}.
\newblock \emph{JMLR}, pages 3371--3408, 2010.

\end{thebibliography}

\end{document}